\documentclass[sigconf]{acmart}
\usepackage{amsmath}
\usepackage{graphicx}
\usepackage{url}
\usepackage{hyperref}
\usepackage{subfig}
\usepackage{amsfonts}  
\usepackage{color, soul}
\usepackage{mathrsfs}
\usepackage{cleveref}
\usepackage{multirow}
\usepackage{wrapfig}
\usepackage{amsthm}
\usepackage{bm}

\AtBeginDocument{%
  \providecommand\BibTeX{{%
    \normalfont B\kern-0.5em{\scshape i\kern-0.25em b}\kern-0.8em\TeX}}}

\copyrightyear{2021} 
\acmYear{2021} 
\setcopyright{acmcopyright}\acmConference[CIKM '21]{Proceedings of the 30th ACM International Conference on Information and Knowledge Management}{November 1--5, 2021}{Virtual Event, QLD, Australia}
\acmBooktitle{Proceedings of the 30th ACM International Conference on Information and Knowledge Management (CIKM '21), November 1--5, 2021, Virtual Event, QLD, Australia}
\acmPrice{15.00}
\acmDOI{10.1145/3459637.3482226}
\acmISBN{978-1-4503-8446-9/21/11}

\begin{document}
\fancyhead{}
\title{AdaGNN: Graph Neural Networks with\\ Adaptive Frequency Response Filter}


\author{Yushun Dong}
\affiliation{%
  \institution{University of Virginia}
    \city{Charlottesville}
  \country{USA}
}
\email{yd6eb@virginia.edu}

\author{Kaize Ding}
\affiliation{%
  \institution{Arizona State University}
      \city{Tempe}
  \country{USA}
}
  \email{kding9@asu.edu}
  
\author{Brian Jalaian}
\affiliation{%
 \institution{Army Research Laboratory}
     \city{Adelphi}
  \country{USA}
 }
 \email{brian.a.jalaian.civ@mail.mil}
 
\author{Shuiwang Ji}
\affiliation{%
  \institution{Texas A$\&$M University}
      \city{College Station}
  \country{USA}
  }
\email{sji@tamu.edu}

\author{Jundong Li}
\affiliation{%
  \institution{University of Virginia}
      \city{Charlottesville}
  \country{USA}
}
\email{jundong@virginia.edu}




\begin{abstract}
Graph Neural Networks have recently become a prevailing paradigm for various high-impact graph analytical problems. Existing efforts can be mainly categorized as spectral-based and spatial-based methods. The major challenge for the former is to find an appropriate graph filter to distill discriminative information from input signals for learning. Recently, myriads of explorations are made to achieve better graph filters, e.g., Graph Convolutional Network (GCN), which leverages Chebyshev polynomial truncation to seek an approximation of graph filters and bridge these two families of methods. Nevertheless, it has been shown in recent studies that GCN and its variants are essentially employing fixed low-pass filters to perform information denoising. Thus their learning capability is rather limited and may over-smooth node representations at deeper layers. To tackle these problems, we develop a novel graph neural network framework AdaGNN with a well-designed adaptive frequency response filter. At its core, AdaGNN leverages a simple but elegant trainable filter that spans across multiple layers to capture the varying importance of different frequency components for node representation learning. The inherent differences among different feature channels are also well captured by the filter. As such, it empowers AdaGNN with stronger expressiveness and naturally alleviates the over-smoothing problem. We empirically validate the effectiveness of the proposed framework on various benchmark datasets. Theoretical analysis is also provided to show the superiority of the proposed AdaGNN. The open-source implementation of AdaGNN can be found here: https://github.com/yushundong/AdaGNN.
\end{abstract}



\maketitle
\section{Introduction}
Graph Neural Networks (GNNs) have demonstrated remarkable performance in a wide spectrum of graph learning tasks, e.g., node classification~\cite{kipf2016semi, hamilton2017inductive,xu2018powerful}, link prediction~\cite{pan2018adversarially, zou2019encoding,kipf2016variational}, and recommendation~\cite{fan2019graph,wang2019neural,fan2019graph}. The main intuition of GNNs is that they stack multiple layers of neural network primitives to learn high-level node feature representations, aiming at addressing various learning tasks in an end-to-end manner~\cite{dong2021individual}. GNNs are deeply influenced by the architecture design of convolutional neural networks (CNNs) for grid-like data such as images and texts~\cite{lecun1998gradient,kim2014convolutional,zhang2015character,he2016deep} and are extended to handle non-Euclidean graph data. In essence, existing GNNs are mainly divided into two main streams: spectral-based and spatial-based methods~\cite{wu2020comprehensive}. With deep roots in Graph Signal Processing (GSP)~\cite{shuman2013emerging} and Spectral Graph Theory~\cite{chung1997spectral}, spectral-based methods~\cite{kipf2016semi,defferrard2016convolutional,levie2018cayleynets,bruna2013spectral} define convolution operations in the spectral domain based on graph Fourier transform and thus bear a solid mathematical foundation. For spatial-based methods, the convolution operations are defined in the spatial domain and act as a message-passing process~\cite{atwood2016diffusion,niepert2016learning,gilmer2017neural}. Specifically, for each node, the convolution operations aggregate and transform information from its neighborhoods when learning its feature representation.

The seminal work of Graph Convolutional Network (GCN)~\cite{kipf2016semi} bridges the gap between these two families of algorithms. As a localized first-order approximation of spectral graph convolution~\cite{defferrard2016convolutional}, GCN can also be interpreted as a spatial-based method with a clear meaning of node localization, thus inspiring a lot of follow-up improvements~\cite{hamilton2017inductive,levie2018cayleynets,atwood2016diffusion,gilmer2017neural,dong2021edits,velivckovic2017graph,xu2018powerful,li2018adaptive,zhuang2018dual,wu2019simplifying}, especially in the spatial domain. However, the fundamental studies and improvements of GCN from the spectral perspective is rather limited. Until fairly recently, studies have shown that the frequency response of the convolution operation corresponds to a fixed low-pass filter at each layer~\cite{nt2019revisiting}, implying that more information is captured within low frequencies and the effects of high-frequency components are much more weakened~\cite{wu2019simplifying}. Despite that, real-world graphs are much more complex than we can imagine, and mixture of different frequencies may either benefit or degrade the final performance. Recent studies have shown that the lowest frequency does not necessarily contain the most important information; while high-frequency components may also encode useful information that is beneficial for the performance under certain tasks~\cite{bo2021beyond,chen2019drop}. In this regard, simply using a fixed low-pass filter cannot well capture the varying importance of different frequency components, thus limiting the expressiveness of learned representations and yielding suboptimal learning performance. Additionally, we show in this paper that in the limit, any filter satisfying certain conditions leads to another fundamental limitation of GCN and its variants -- the over-smoothing problem~\cite{li2018deeper,chen2020measuring,cai2020note,liu2020towards}. It refers to the phenomenon that node feature representations converge to similar values at deeper layers, thus nodes cannot be easily distinguished. 


To tackle the aforementioned problems, we overcome the limitations of GCN and its variants from the spectral perspective. Specifically, we propose a novel framework AdaGNN with an adaptive frequency response filter, which adaptively adjusts the importance of different frequency components for spectral convolution when multiple layers are stacked. It should be noted that a straightforward solution to control the varying effects of different frequency components is allocating a learnable parameter for each frequency component. However, this strategy requires expensive eigendecomposition~\cite{bruna2013spectral} and its large complexity makes the model prone to overfitting especially when training data is limited. Instead, we propose a simple but elegant solution to assign a single parameter for each feature channel at each layer, based upon which we stack multiple layers to learn a flexible and powerful filter. In a nutshell, the main contributions of this paper can be summarized as follows:

\begin{itemize}
    \item We systematically examine the fundamental limitations of fixed low-pass filters of GCN and its variants from the perspective of spectral domain.
    
    \item We develop a novel graph neural network framework named AdaGNN that can capture the varying importance of different frequency components for node representation learning. The core of this framework is a simple but elegant trainable filter that spans across multiple layers with a single parameter for each feature channel at each layer. The developed GNN framework does not involve expensive eigendecomposition and its parameter complexity is comparable to the lightweight GCN model such as SGC~\cite{wu2019simplifying}.
    
    \item We provide theoretical analysis for the proposed framework. Firstly, we show that the filter of prevalent GCN models (e.g., GCN and mean aggregator of GraphSage~\cite{hamilton2017inductive}) can be considered as a special case of our proposed adaptive filter at each layer. Secondly, we provide both spectral and spatial interpretation of the proposed framework. Thirdly, we also prove that any filter satisfying certain conditions will inevitably encounter over-smoothness in deeper structure, while the proposed filter can naturally alleviate this issue. 
    
    \item We conduct comprehensive experiments on benchmark graph datasets with different properties. The empirical evaluations demonstrate that the proposed AdaGNN not only learns more powerful node representations for the node classification task but also greatly alleviates the over-smoothing problems when the architecture goes deeper. 
\end{itemize}

\section{The Proposed Framework -- AdaGNN}

Here we firstly introduce the notations and other commonly used preliminaries in this paper. Then we introduce details about how we develop the graph filter with adaptive frequency response filtering. Finally, we present the overall architecture of AdaGNN.

\subsection{Notations and Preliminaries} 
\textbf{Notations.} We define an undirected graph as $G = (\mathcal{V}, \mathcal{E})$, where $\mathcal{V}=\{v_{1},...,v_{N}\}$ and $\mathcal{E}$ denote the set of nodes and edges, respectively. Let $\mathbf{A}\in\mathbb{R}^{N \times N}$ be the adjacency matrix of the graph such that $\mathbf{A}_{i,j} = 1$ if $v_{j} \in \mathcal{N}(v_{i})$, otherwise $\mathbf{A}_{i,j} = 0$. Here $\mathcal{N}(.)$ denotes the one-hop neighbor set of a node. The Laplacian matrix of the graph is 
defined as $\mathbf{L} = \mathbf{D}-\mathbf{A}$, where $\mathbf{D} = \text{diag}(d_{1}, ...,d_{N})$ is the diagonal degree matrix ($d_{i} = \sum_{j} \mathbf{A}_{i, j}$). Then the symmetric normalized Laplacian matrix and the random-walk normalized Laplacian matrix are defined as $\mathbf{L}_{sym}=\mathbf{D}^{-\frac{1}{2}}\mathbf{L}\mathbf{D}^{-\frac{1}{2}}$ and $\mathbf{L}_{rw}=\mathbf{D}^{-1}\mathbf{L}$, respectively. Besides, feature matrix $\mathbf{X}\in \mathbb{R}^{N \times F}$ is utilized to describe properties of nodes, where $\mathbf{x}_j$ (column of $\mathbf{X}$) represents $j$-th feature channel of $\mathbf{X}$ and $F$ denotes the number of feature channels.

\textbf{Graph Filters.} The main idea of spectral-based GNN methods is to define graph filters based on Graph Signal Processing~\cite{shuman2013emerging}. Specifically, symmetric normalized Laplacian can be factored as $\mathbf{L}_{sym}=\mathbf{D}^{-\frac{1}{2}}\mathbf{L}\mathbf{D}^{-\frac{1}{2}} = \mathbf{U} \mathbf{\Lambda} \mathbf{U}^{T}$. Here $\mathbf{U} \in \mathbb{R}^{N \times N} = [\mathbf{u}_1,...,\mathbf{u}_N]$, where $\mathbf{u}_i\in\mathbb{R}^{N}$ denotes the $i$-th eigenvector of $\mathbf{L}_{sym}$ and $\mathbf{\Lambda} = \text{diag}(\lambda_1,, ..., \lambda_N)$ is the corresponding eigenvalue matrix. Let $\mathbf{x}\in\mathbb{R}^{N}$ be an one-channeled input signal of all nodes, then the Graph Fourier transform and inverse Fourier transform can be defined as $\hat{\mathbf{x}} = \mathscr{F}(\mathbf{x}) = \mathbf{U}^{T}\mathbf{x}$ and $\mathbf{x} = \mathscr{F}^{-1}( \hat{\mathbf{x}}) = \mathbf{U} \hat{\mathbf{x}}$, respectively. Here $\hat{\mathbf{x}}$ is the Fourier transformed graph signal. Graph convolution of the input signal $\mathbf{x}$ with filter $\mathbf{g} = \text{diag(} \bm{\theta} \text{)}$ parameterized by $\bm{\theta} \in \mathbb{R}^{N}$ is defined as $\mathbf{x} *_{G} \mathbf{g} = \mathbf{U} \mathbf{g}(\mathbf{\Lambda}) \mathbf{U}^{T} \mathbf{x}$. A vast majority of existing works such as GCN and SGC~\cite{kipf2016semi,wu2019simplifying} use a fixed low-pass filter for the graph convolution operation while recent studies~\cite{wu2019simplifying,nt2019revisiting} have shown that if the input signal is repeatedly convolved with the fixed low-pass filter, its high-frequency components will be greatly weakened and the learning performance is dominated by the low-frequency components, resulting in the well-known over-smoothing problem.

\begin{figure*}[!t]
	\centering  
	\includegraphics[width=0.90\linewidth]{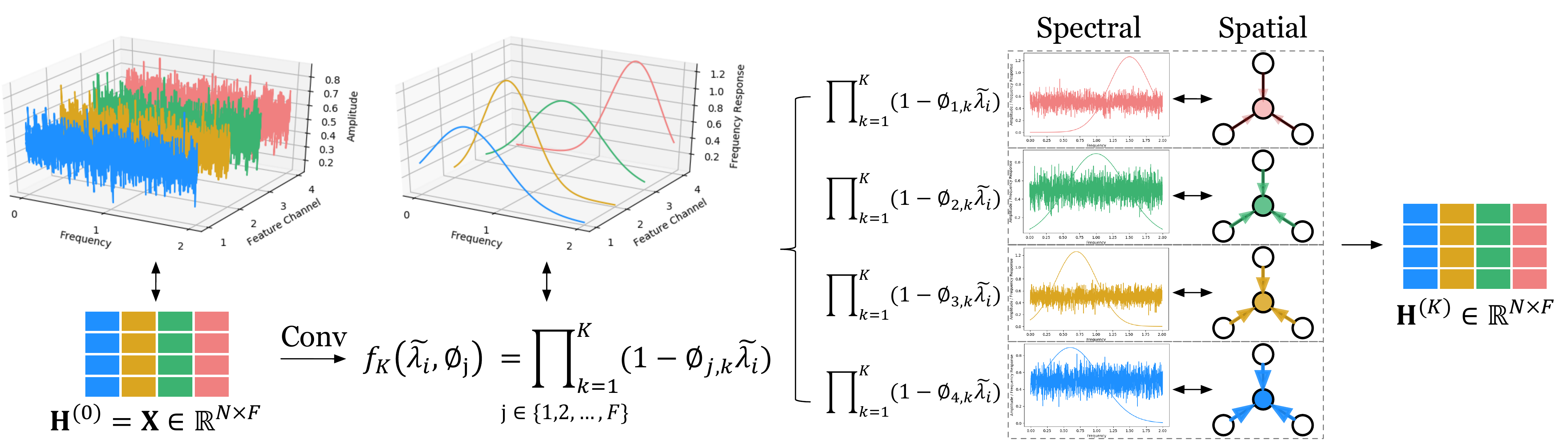} 
	\vspace{1mm}
	\caption{An illustration of the proposed AdaGNN from the spectral perspective. The convolution operation across $K$ layers is equivalent to applying the filter $\prod_{k=1}^{K} (1- \phi_{j, k} \tilde{\lambda_{i}})$ to the $j$-th channel of input signal $\mathbf{x}_{j}$. Here we omit the weight matrix $\mathbf{\Theta}$ and ReLU function at the first layer for the ease of understanding.}  
	\label{fig:adagnn}   
\vspace{2mm}
\end{figure*}

\subsection{Adaptive Frequency Response Filtering}
In image signal processing, the Laplacian kernel is widely used to capture high-frequency edge information for various tasks such as image sharpening and blurring~\cite{gonzales2002digital, he2010guided}. As its counterpart in GSP, we can multiply the graph Laplacian matrix $\mathbf{L}$ with the input graph signal $\mathbf{x}\in\mathbb{R}^{N}$ (i.e., $\mathbf{h}=\mathbf{L}\mathbf{x}$) to characterize its high-frequency components -- the frequencies that carry sharply varying signal information across edges of graph. Meanwhile, as shown in recent studies~\cite{wu2019simplifying,yang2016revisiting}, the essence of GCN and its variants are the low-pass filter which smoothes the feature representations of the current node and its neighbors to make them similar. As such, we can highlight the low-frequency components of input signal $\mathbf{x}$ by setting $\mathbf{z}=\mathbf{x}-\mathbf{L}\mathbf{x}$, i.e., subtracting the term $\mathbf{L}\mathbf{x}$ which emphasizes more on high-frequency components from the input signal $\mathbf{x}$. In fact, this formulation is well aligned with the convolution operation in GCN~\cite{kipf2016semi} if we replace $\mathbf{L}$ with the symmetric normalized Laplacian matrix $\tilde{\mathbf{L}}_{sym}$ which is derived from the self-loop augmented adjacency matrix $\tilde{\mathbf{A}}$:
\vspace{-0.05in}
\begin{equation}
    \mathbf{z}=\mathbf{x}-\tilde{\mathbf{L}}_{sym}\mathbf{x}=(\mathbf{I}-\tilde{\mathbf{D}}^{-\frac{1}{2}}(\tilde{\mathbf{D}}-\tilde{\mathbf{A}})\tilde{\mathbf{D}}^{-\frac{1}{2}})\mathbf{x}= \tilde{\mathbf{D}}^{-\frac{1}{2}}\tilde{\mathbf{A}}\tilde{\mathbf{D}}^{-\frac{1}{2}}\mathbf{x},
\end{equation}
where $\tilde{\mathbf{A}}=\mathbf{A}+\mathbf{I}$ is used to avoid numerical instability issues~\cite{kipf2016semi} and $\tilde{\mathbf{D}}$ is the degree matrix from $\tilde{\mathbf{A}}$. Other than that, we can also use the random-walk normalized Laplacian matrix $\tilde{\mathbf{L}}_{rw}$ from $\tilde{\mathbf{A}}$. Here, we use $\tilde{\mathbf{L}}$ to denote the general formulation and it can be instantiated as $\tilde{\mathbf{L}}_{sym}$ and $\tilde{\mathbf{L}}_{rw}$.

It should be noted that the above operation corresponds to a fixed low-pass filter in the spectral domain, where higher weights are specified for low-frequency components. However, in practice, low-frequency components may not always be useful, and high-frequency can also provide complementary insights for learning~\cite{bo2021beyond,chen2019drop}, especially when the label information is not smooth across edges. Additionally, at deeper layers~\cite{wu2019simplifying}, the high-frequency components of the input graph signal are unavoidably too much weakened compared with the lower ones with fixed filters, leading to the well-known over-smoothing problem~\cite{li2018deeper}. In this regard, the fixed low-pass filters largely limit the fitting capability of GCN and its variants for learning discriminative node representations. As a consequence, it is vital to capture the varying importance of frequencies in the filter to preserve more useful information and alleviate over-smoothing in deeper layers.

As an alternative of the traditional fixed filter, we propose a novel adaptive frequency response filter to tackle the aforementioned problems. The developed filter should be learnable and able to adaptively adjust the varying importance of different frequencies for convolution when multiple layers are stacked together. Toward this goal, one straightforward solution is to assign a learnable parameter for each frequency component at each layer to increase fitting capability. Nonetheless, this solution requires explicit eigendecomposition which is too expensive. Our solution to address these issues is simple but elegant -- we assign a single parameter per feature dimension for the low-pass filter at each layer  and a more powerful filter can then be built when multiple layers are stacked together. Specifically, the filter at each layer is formulated as $\mathbf{z}_j=\mathbf{x}_j-\phi\tilde{\mathbf{L}}\mathbf{x}_j$ ($1 \leq j \leq F$) for the $j$-th feature channel, where $\phi$ is a learnable parameter. We can also generalize it to a multi-channeled input signal $\mathbf{X}\in\mathbb{R}^{N\times F}$ that has $F$ different feature channels:
\begin{equation}
    \mathbf{E}=\mathbf{X}-\tilde{\mathbf{L}}\mathbf{X}\mathbf{\Phi},
    \label{eq:filtering}
\end{equation}
where $\mathbf{\Phi}=\text{diag}(\phi_1,...,\phi_{F})$, and $\phi_j$ denotes the learnable parameter for the $j$-th feature channel. $\mathbf{E}$ denotes the node representation matrix after the filtering operation. Intuitively, this channel-specific learnable parameter helps to achieve how much high-frequency component should be weakened. Theoretical analysis on how such a simple operation enables to learn a more flexible filter and naturally helps alleviate over-smoothing is presented in Section \ref{theoretical_analysis}.

\subsection{Overall Architecture of AdaGNN}
In this subsection, we mainly introduce the overall architecture of our proposed AdaGNN, which is mainly composed of two different components: node representation learning and label prediction.

\textbf{Node Representation Learning.} Based on the above discussion, we can stack $K$ layers of convolution operations as Eq.~(\ref{eq:filtering}) for a more powerful filter. Traditionally, each layer takes the output of the previous layer as input and is transformed with a weight matrix followed by a nonlinear activation function. Inspired by~\cite{wu2019simplifying}, we can remove the noncritical nonlinear activation functions and weight matrices at intermediate layers and only keep those at the first layer. The reason is two-fold: (1) nonlinear feature transformation that may benefit learning 
is still preserved; (2) the number of model parameters  greatly decreases by reducing the input feature channels for the second and later layers. Specifically, suppose $\mathbf{H}^{(0)}=\mathbf{X}$, then we have $\mathbf{H}^{(1)}=\text{ReLU}(\mathbf{E}\mathbf{\Theta})=\text{ReLU}((\mathbf{H}^{(0)}-\tilde{\mathbf{L}}\mathbf{H}^{(0)}\mathbf{\Phi}_{1})\mathbf{\Theta})$ as the output of the first layer, where $\mathbf{\Phi}_{1}$ (diagonal matrix) and $\mathbf{\Theta}\in\mathbb{R}^{F\times L}$ are the learnable parameters for the filter and the weight matrix at the first layer, respectively. As mentioned before, we often set $L<F$. For the intermediate layer $2\leq k\leq K$, we have $\mathbf{H}^{(k)}=\mathbf{H}^{(k-1)}-\tilde{\mathbf{L}}\mathbf{H}^{(k-1)}\mathbf{\Phi}_{k}$, where $\mathbf{\Phi}_{k}$ is the learnable parameters for the $k$-th layer. The output representation after $K$ layers are $\mathbf{H}^{(K)}\in\mathbb{R}^{N\times L}$. A learning illustration from the spectral perspective is shown in Fig.~\ref{fig:adagnn}. According to the theoretical analysis in Section \ref{theoretical_analysis}, each feature channel has its own filter, whose frequency response function can be adaptively learned to capture the useful information in different frequency components. Due to the filters for different feature channels are decoupled from each other, appropriate levels of smoothness can be individually achieved for each feature channel, which provides strengthened fitting ability.

\textbf{Label Prediction.} At the final layer (i.e., the $K$-th layer) of AdaGNN, the output $\mathbf{H}^{(K)}$ is further fed into a softmax classifier to obtain the probability of nodes in $C$ different classes. In particular, we have $\hat{\mathbf{Y}}=\text{softmax}(\mathbf{H}^{(K)}\mathbf{W})$, where $\mathbf{W}\in\mathbb{R}^{L\times C}$ is a weight matrix to transform node representations to the label space. A straightforward choice for the loss function of classification task is cross-entropy loss. Besides, regularizations are also considered for the proposed AdaGNN to avoid over-fitting. Then the loss function of the proposed AdaGNN framework is formulated as follows:
\begin{align} 
\label{loss}
\mathcal{L} = -\sum_{i\in\mathcal{Y}_L}   \sum_{j=1}^{C}\mathbf{Y}_{ij} \text{ln}\hat{\mathbf{Y}}_{ij} &+ \alpha \sum_{k=1}^{K} \|\mathbf{\Phi}_{k}\|_{1} \notag \\
&+\beta(\sum_{k=1}^{K} \|\mathbf{\Phi}_k\|_{F}^{2}+\|\mathbf{\Theta}\|_{F}^{2}+\|\mathbf{W}\|_{F}^{2}),
\end{align} 
where $\mathcal{Y}_L$ is the set of labeled nodes indices, and $\mathbf{Y}\in\mathbb{R}^{|\mathcal{Y}_L|\times C}$ denotes the ground truth labels. $\alpha$ and $\beta$ here are hyper-parameters. The first term is the cross-entropy loss between predictions and ground truth on labeled nodes. The second term is the $\ell_1$-norm regularization of $\mathbf{\Phi}_k$ for sparsity. The third term is the $\ell_{2}$-norm regularization for all trainable parameters to prevent overfitting. The effect of the two regularization terms in Eq. (\ref{loss}) are controlled by tuning $\alpha$ and $\beta$, respectively.

\begin{figure*}[!t]
	\centering 
	\vspace{3pt}
	\includegraphics[width=0.98\linewidth]{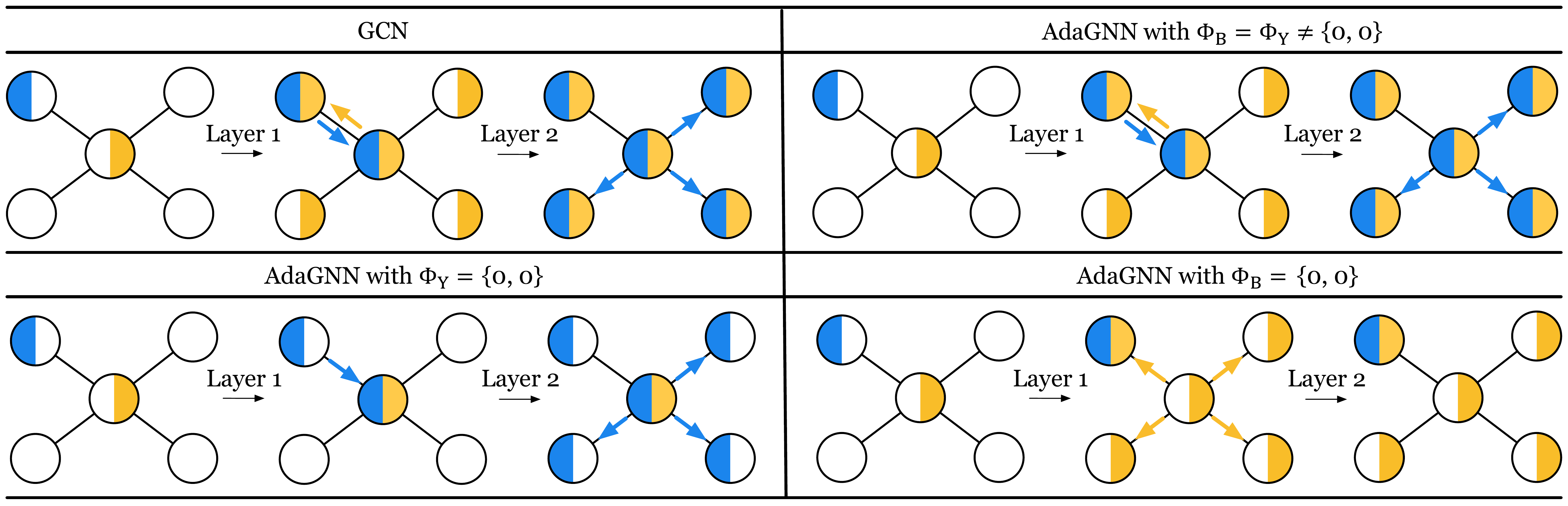}
	\vspace{1pt}
	\caption{An illustrative example to show how the learnable parameters of AdaGNN controlling the smoothness of different feature channels benefits the model fitting ability.}  
	\label{fig:toy_example}   
\end{figure*}

\section{Theoretical Analysis}
In this section, we firstly introduce the connections of AdaGNN to GCN and GraphSAGE to gain a deeper understanding of the essence of our proposed framework. Then we carry out spectral analysis for AdaGNN to demonstrate its frequency response function. Besides, spatial analysis of AdaGNN is presented to illustrate our proposed graph filter from the spatial perspective. Finally, theoretical analysis on over-smoothing demonstrate how our proposed framework naturally helps to alleviate over-smoothing.

\label{theoretical_analysis}
\subsection{Connections to GCN and GraphSAGE}
\label{special_case}
\textbf{Observation 1.} \textit{The aggregation operation of GCN reduces to the operation of AdaGNN defined in Eq.~(\ref{eq:filtering}) when $\tilde{\mathbf{L}}=\tilde{\mathbf{L}}_{sym}$ and $\mathbf{\Phi}=\mathbf{I}$; GraphSAGE aggregation operation with mean aggregator and sampling rate being 1 is also a special case when $\tilde{\mathbf{L}}=\tilde{\mathbf{L}}_{rw}$ and $\mathbf{\Phi}=\mathbf{I}$.}

\begin{proof}
The information aggregation of AdaGNN with $\tilde{\mathbf{L}}_{sym}$ and $\tilde{\mathbf{L}}_{rw}$ at layer $k$ can be respectively formulated as follows:
\begin{align} 
\mathbf{E}^{(k)}_{v,j} & = \mathbf{H}^{(k)}_{v,j} - \phi_{j, k} \sum_{u \in \mathcal{N}(v)} (\tilde{\mathbf{L}}_{sym})_{v,u} \mathbf{H}^{(k)}_{u,j}, \;\;\; \mbox{and}\\
\mathbf{E}^{(k)}_{v,j} &= \mathbf{H}^{(k)}_{v,j} - \phi_{j, k} \sum_{u \in \mathcal{N}(v)} (\tilde{\mathbf{L}}_{rw})_{v,u} \mathbf{H}^{(k)}_{u,j},
\end{align} 

Considering that
\begin{align}
(\tilde{\mathbf{L}}_{sym})_{v,u}&=\left\{
\begin{array}{cl}
-\frac{1}{\sqrt{|\mathcal{N}(v)|+1}\sqrt{|N(u)|+1}}, &\mbox{if}\;\; u \in \mathcal{N}(v), \\
\;\;\;\;\;\frac{|\mathcal{N}(v)|}{|\mathcal{N}(v)|+1}, &\mbox{if}\;\; u =v,  \\
\;\;\;\;0,  &\mbox{otherwise},
\end{array}
\right. \,\,\, \mbox{and}  \notag\\
(\tilde{\mathbf{L}}_{rw})_{v,u}&=\left\{
\begin{array}{cl}
\;\;\;\;\;\;\;\;\;-\frac{1}{|\mathcal{N}(v)|+1},  \;\;\;\;\;\;\;\;\; &\mbox{if}\;\;u \in \mathcal{N}(v), \\
\;\;\;\;\;\;\;\;\;\;\;\;\frac{|\mathcal{N}(v)|}{|\mathcal{N}(v)|+1},  \;\;\;\;\;\;\;\;\;  &\mbox{if}\;\;u =v,  \\
\;\;\;\;\;\;\;\;\;\;\;\;0,  \;\;\;\;\;\;\;\;\; &\mbox{otherwise},
\end{array}
\right.\notag
\end{align}
we have the following two formulations:
\begin{align}
\mathbf{E}^{(k)}_{v,j} 
\label{sym_gcn}
&= (1-(1 - \frac{1}{|\mathcal{N}(v)|+1})\phi_{j,k})\mathbf{H}^{(k)}_{v,j}  \notag \\
&\;\;\;\;\;\;\;\;\;\;\;\;\;\;\; + \phi_{j,k} \sum_{u \in \mathcal{N}(v)} \frac{\mathbf{H}^{(k)}_{u,j}}{\sqrt{|\mathcal{N}(v)|+1} \sqrt{|\mathcal{N}(u)|+1}},   \text{ and} \\
\mathbf{E}^{(k)}_{v,j} 
\label{rw_sage}
&=(1-(1 - \frac{1}{|\mathcal{N}(v)|+1})\phi_{j,k})\mathbf{H}^{(k)}_{v,j} + \phi_{j,k}\sum_{u \in \mathcal{N}(v)} \frac{\mathbf{H}^{(k)}_{u,j}}{|\mathcal{N}(v)|+1} .
\end{align}
Then if we replace all $\phi_{j,k}$ by 1, and Eq.~(\ref{sym_gcn}) and Eq.~(\ref{rw_sage}) can be respectively reformulated as
\begin{align}
\mathbf{E}^{(k)}_{v,j} &= \frac{1}{|\mathcal{N}(v)| + 1}\mathbf{H}^{(k)}_{v,j} + \sum_{u \in \mathcal{N}(v)} \frac{\mathbf{H}^{(k)}_{u,j}}{\sqrt{|\mathcal{N}(v)|+1} \sqrt{|\mathcal{N}(u)|+1}}    , \text{  and} \\
\mathbf{E}^{(k)}_{v,j} &= \frac{1}{|\mathcal{N}(v)|+1}\mathbf{H}^{(k)}_{v,j} + \sum_{u \in \mathcal{N}(v)} \frac{\mathbf{H}^{(k)}_{u,j}}{|\mathcal{N}(v)|+1} . \,\,\,
\end{align}
Therefore, the convolution operation of AdaGNN with $\tilde{\mathbf{L}}_{sym}$ and $\mathbf{\Phi} = \mathbf{I}$ equals to that of GCN. Also, the convolution operation of AdaGNN with $\tilde{\mathbf{L}}_{rw}$ and $\mathbf{\Phi} = \mathbf{I}$ equals to that of GraphSAGE with mean aggregator and sampling rate being 1.
\end{proof}

Observation 1 reveals the inherent connections between AdaGNN and prevalent GNNs such as GCN~\cite{kipf2016semi} and GraphSAGE~\cite{hamilton2017inductive}.  As the aggregation operations of GCN and GraphSAGE are fundamental building blocks of modern GNN architectures~\cite{gao2019graph,ying2018hierarchical}, it shows the broad generalization of our proposed framework. 

\subsection{Spectral Analysis of AdaGNN}
\label{spectral}
Here we provide a formal spectral analysis of the frequency response of AdaGNN framework with $K$ layers. Spectral analysis is based on  $\tilde{\mathbf{L}}_{sym}$ and here we omit the weight matrix and activation function in the first layer for ease of analysis~\cite{wu2019simplifying}.

\textbf{Theorem 1.} \textit{For a K-layer AdaGNN framework, its frequency response function of the $j$-th input feature channel is formulated as $f_K(\tilde{\lambda}_i, \phi_j) = \prod_{k=1}^{K} g_k(\tilde{\lambda}_i, \phi_{j,k}) = \prod_{k=1}^{K} (1- \phi_{j,k} \tilde{\lambda}_{i})$, where $\phi_{j,k}$ denotes the learnable parameter of $j$-th feature channel at layer $k$.}

\begin{proof}
As mentioned in Section~\ref{spectral}, we replace $\tilde{\mathbf{L}}$ with $\tilde{\mathbf{L}}_{sym}$ and omit the weight matrix and activation function for the frequency response function proof. Consider the $j$-th feature channel $\mathbf{x}_j \in \mathbb{R}^{N}$ of the input signal at the $k$-th layer, then we have
\begin{equation*} 
\begin{aligned}
\mathbf{x}_j - \phi_{j,k}\tilde{\mathbf{L}}_{sym}\mathbf{x}_j &= \mathbf{x}_j - \phi_{j,k}\mathbf{\tilde{U}}\mathbf{\tilde{\Lambda}}\mathbf{\tilde{U}}^{T}\mathbf{x}_{j}  \\
&= (\mathbf{\tilde{U}} \mathbf{\tilde{U}}^{T} - \phi_{j,k}\mathbf{\tilde{U}}\mathbf{\tilde{\Lambda}}\mathbf{\tilde{U}}^{T})\mathbf{x}_j  \\
&= \mathbf{\tilde{U}} (\mathbf{I} - \phi_{j,k} \tilde{\mathbf{\Lambda}}) \mathbf{\tilde{U}}^{T} \mathbf{x}_j.
\end{aligned}
\end{equation*} 
As a consequence, the frequency response of AdaGNN can be derived as $g_{k}(\tilde{\lambda}_i) = 1-\phi_{j,k}\tilde{\lambda}_i$. If we stack $K$ layers together, then the overall frequency response function can be derived as $f_{K}(\tilde{\lambda}_i, \phi_j) = \prod_{k=1}^{K}g_{k}(\tilde{\lambda}_i, \phi_{j,k}) = \prod_{k=1}^{K} (1- \phi_{j,k} \tilde{\lambda}_{i})$.
\end{proof}

Compared with the $K$-layered GCN whose frequency response function is $f_K(\tilde{\lambda}_i) = (1- \tilde{\lambda}_{i})^K$ for any feature channel\footnote{Here we also omit the weight matrix and nonlinear transformation function at each layer for fair comparison.}, the advantages of AdaGNN are three-fold: (1) the parameter $\phi_{j,k}$ adjusts the relative importance of high-frequency and low-frequency components at each layer $k$; (2) when multiple layers are stacked, the importance of different frequency components can be better captured as a set of trainable parameters $\{\phi_{j,1},...,\phi_{j,K}\}$, yielding a more complex frequency response function; (3) the frequency response function of each feature channel is decoupled from each other, providing them with more flexibility to achieve different levels of smoothness and to learn more discriminative node representations.

\subsection{Spatial Analysis of AdaGNN}
\label{spatial}
\textbf{Corollary 1.} \textit{AdaGNN adaptively adjusts the smoothness of each feature channel via learnable parameter $\bm{\Phi}$ in the information aggregation process in the spatial domain.}

Based on Obervation 1, we learn that the weights in $\bm{\Phi}$ can be regarded as a smoothness controller in the information aggregation process at each layer. When the corresponding weights in $\phi$ is large for a certain feature channel, then more information will be aggregated to the central node in this channel; however, when when the corresponding weights in $\phi$ is small for a certain feature channel, less information will flow into a node from its neighbors. In this regard, the proof of Corollary 1 is straightforward, i.e., AdaGNN adaptively adjusts the smoothness of eachfeature channel via controlling how much information can flow into a node from its neighbors. To better illustrate how this process helps achieve better information propagation and learn more discriminative node representations in the spatial domain, we provide a toy example with different exemplary operators, which is introduced as follows.

\textbf{An illustrative Example.} In Fig.~\ref{fig:toy_example}, we provide an illustrative example to show how different feature channels learn different levels of smoothness, which is essential to alleviate the over-smoothing problem. Here we have five nodes and two feature channels (blue and yellow). At the very beginning, the upper left node is associated with the blue channel and the middle node is with the yellow channel. For GCN, features propagate with the same mechanism and 
the node representations become the same after two layers. For AdaGNN, suppose the learnable parameters for these two channels across two layers are $\Phi_{B}=\{\phi_{B,1},\phi_{B,2}\}$ and $\Phi_{Y}=\{\phi_{Y,1},\phi_{Y,2}\}$. When the parameters for these two channels are the same, AdaGNN still suffers from the over-smoothing problem (upper right subfigure). By adaptively learning the optimal parameters, we can easily control the smoothness of each feature channel, which naturally alleviates the over-smoothing problem (the lower two subfigures).

\subsection{Over-smoothing Analysis}
\label{oversmoothness}
Here we show why over-smoothing is inevitable in GCN and its variants with fixed low-pass filters, and why our proposed AdaGNN can naturally alleviate the over-smoothing problem.

\textbf{Theorem 2.} \textit{For any fixed low-pass filters defined over $\tilde{\mathbf{L}}_{sym}$, we assume $\tilde{\lambda}_{1}$ is the smallest eigenvalue. Given a graph signal $\mathbf{x}$, suppose we convolve $\mathbf{x}$ with the filter across $K$ layers (assume the filter is $g_{k}(.)$ at layer $k$). If the total frequency response satisfies that $\lim \limits_{K \to \infty} \prod_{k=1}^{K} g_{k}(\tilde{\lambda}_{i}) = 0$ ($\forall \tilde{\lambda}_i\neq \tilde{\lambda}_1$), then over-smoothing issue is inevitable (i.e., feature values of different nodes become the same), and vice versa.}

\begin{proof}
Consider a graph signal $\mathbf{x} \in \mathbb{R}^N$ as input. Here we assume that the value of at least one dimension in $\mathbf{x}$ is different from other dimensions, i.e., $\mathbf{x}$ is not an over-smoothed signal and $N\geq 2$.

\textbf{Fact 1.} \textit{For any $\tilde{\mathbf{L}}_{sym} = \tilde{\mathbf{U}} \tilde{\mathbf{\Lambda}} \tilde{\mathbf{U}}^{T} \in \mathbb{R}^{N \times N}$ of an undirected graph, columns in $\tilde{\mathbf{U}}$ are orthogonal to each other, and eigenvector $\mathbf{\tilde{u}}_{1}$ corresponding to the smallest eigenvalue $\tilde{\lambda}_{1}$ is collinear with the vector $[1,1,1,...,1]$ of length $N$.}

As such, we can regard $\tilde{\mathbf{L}}_{sym} \mathbf{x} = \tilde{\mathbf{U}} \tilde{\mathbf{\Lambda}} \tilde{\mathbf{U}}^{T}\mathbf{x}$ as a process of projecting $\mathbf{x} $ onto $N$ eigenvectors, then re-weighting the length of each component vector and summing them together. Assume weight of each frequency component (i.e., each $\tilde{\lambda}_i$, $1 \leq i \leq N$) of the filter defined over $\tilde{\mathbf{L}}_{sym}$ at filtering time $k$ ($1 \leq k \leq K$) being $g_k(\tilde{\lambda}_i)$, then $\mathbf{x}_K$, i.e., $\mathbf{x}$ being filtered $K$ times, will be 
\begin{align} 
\mathbf{x}_K = \prod^K_{k=1} g_k(\tilde{\lambda}_1) \frac{\mathbf{u}_1 \cdot \mathbf{x}}{|\mathbf{u}_1|^2} \mathbf{u}_1 &+ \prod^K_{k=1} g_k(\tilde{\lambda}_2) \frac{\mathbf{u}_2 \cdot \mathbf{x}}{|\mathbf{u}_2|^2} \mathbf{u}_2  \\ \notag
&+ ... + \prod^K_{k=1} g_k(\tilde{\lambda}_N) \frac{\mathbf{u}_N \cdot \mathbf{x}}{|\mathbf{u}_N|^2} \mathbf{u}_N.
\end{align} 

\textbf{Fact 2.} \textit{If all entries of an $N$-dimensional nonzero vector are the same, then this vector is collinear with the vector $[1,1,1,...,1]$ of length $N$ (N $\geq 2$).}

Based on Fact 2, we learn that $\mathbf{x}_K$ is collinear with $\mathbf{u}_1$ when $K \rightarrow \infty$ (i.e., over-smoothing issue being inevitble when $K$ gets larger) iff  $\lim \limits_{K \to \infty} \prod_{k=1}^{K} g_{k}(\tilde{\lambda}_{i}) = 0$ ($\forall \tilde{\lambda}_i\neq \tilde{\lambda}_1$).
\end{proof}

As the filter of conventional GCN and its variants are mainly defined over $\tilde{\mathbf{L}}_{sym}$ and satisfy the above condition at extremely deep layers, thus they often suffer from the over-smoothing problem. Meanwhile, we have the following corollary for AdaGNN.

\textbf{Corollary 2.} \textit{Our AdaGNN model can naturally alleviate the over-smoothing problem at deeper layers.}

The proof of Corollary 2 is very straightforward. In Theorem 1, we have shown that for any feature channel $j$, its frequency response function over $K$ layers is $\prod_{k=1}^{K} (1- \phi_{j,k} \tilde{\lambda}_{i})$, where $0 \leq \tilde{\lambda}_{i} < 2$~\cite{wu2019simplifying}. The trainable parameter $\phi_{j,k}$ can help adjust the value of frequency response to ensure it does not satisfy the condition in Theorem 2 (i.e., preventing the total frequency response from approaching 0), naturally alleviating the over-smoothing problem.

\begin{table*}[!htbp]
\centering
\caption{Detailed statistics of the datasets in our experiments.} 
\label{tb:Statistics}
\begin{tabular}{c|c|c|c|c|c|c}
\hline
                 &\textbf{BlogCatalog} &\textbf{Flickr} & \textbf{ACM}    & \textbf{Cora} & \textbf{Citeseer} & \textbf{Pubmed} \\
\hline
\textbf{\# Nodes}                & 5,196     & 7,575     &  16,484               & 2,708   & 3,327     & 19,717      \\
\textbf{\# Edges}                & 173,468   & 242,146   &  71,980                  & 5,429   & 4,732     & 44,338   \\
\textbf{\# Features}              & 8,189     & 12,047    & 8,337             & 1,433   & 3,703     & 500      \\
\textbf{\# Average Degree}               & 66.8      & 63.9      &  8.7               & 4.0     & 2.8       & 4.5            \\
\textbf{\# Classes}                  & 6         & 9         &  9                     & 7       & 6         & 3    \\
\hline
\end{tabular}
\vspace{5pt}
\end{table*}

\section{Experimental Evaluations}
\label{experiments}
In this section, we perform experiments on several real-world datasets to validate the effectiveness of AdaGNN. In particular, we aim to answer the following research questions -- \textbf{RQ1}: How does AdaGNN perform compared with other state-of-the-art spectral GNNs and corresponding variants? \textbf{RQ2}: How well can AdaGNN alleviate the over-smoothing problems at deeper layers? \textbf{RQ3}: In which way will different frequency components contribute to learning? \textbf{RQ4}: How much the adaptive frequency response filter of the proposed AdaGNN contributes to the over-smoothing alleviation?

\begin{table*}[!htbp]
\centering
\caption{Average accuracy with standard deviation on ACM (the performance of AdaGNN-R and AdaGNN-S is marked in \textbf{bold}). Result of GraphSAGE with 128 layers is omitted due to gradient instability in our hyper-parameter search space.}
\label{deeper_results_rearranged2}
\begin{tabular}{c|c|cccc}
\hline
\textbf{Dataset}                      & \textbf{Model}        & \textbf{2 Layer}   & \textbf{8 Layer}   & \textbf{32 Layer}   & \textbf{128 Layer}  \\
\hline
\multirow{8}{*}{ACM}  
                      & GCN          & 75.55 $\pm$ 0.2\% & 73.54 $\pm$ 0.3\% & 53.74 $\pm$ 1.9\% & 35.97 $\pm$ 0.7\% \\
                      & GraphSAGE    & 75.29 $\pm$ 0.2\% & 73.12 $\pm$ 0.6\% & 52.21 $\pm$ 1.5\% & --- \\
                      & SGC          & 73.83 $\pm$ 0.4\% & 75.14 $\pm$ 0.1\% & 73.80 $\pm$ 0.3\% & 64.01 $\pm$ 0.4\% \\
                      & DropEdge-GCN & 73.05 $\pm$ 0.6\% & 72.44 $\pm$ 0.8\% & 71.43 $\pm$ 2.2\% & 67.37 $\pm$ 1.9\% \\
                      & Pairnorm-GCN-SI      & 74.87 $\pm$ 0.3\%      & 74.05 $\pm$ 0.1\%      & 73.63 $\pm$ 0.2\%      & 68.35 $\pm$ 2.0\%     \\
                      & Pairnorm-GCN-SCS     & 75.44 $\pm$ 0.1\%      & 73.01 $\pm$ 0.3\%      & 73.33 $\pm$ 0.1\%      & 70.84 $\pm$ 1.4\%      \\
                      \cline{2-6} 
                      & \textbf{AdaGNN-R}       & \textbf{75.14 $\pm$ 0.3\%} & \textbf{75.27 $\pm$ 0.1\%} & \textbf{75.15 $\pm$ 0.0\%} & \textbf{74.55 $\pm$ 0.1\%} \\
                      & \textbf{AdaGNN-S}       & \textbf{75.65 $\pm$ 0.2\%} & \textbf{75.83 $\pm$ 0.4\%} & \textbf{75.64 $\pm$ 0.0\%} & \textbf{74.95 $\pm$ 0.1\%} \\
\hline
\end{tabular}
\end{table*}

\begin{table*}[!htbp]
\centering
\caption{Average accuracy with standard deviation on BlogCatalog, Flickr, Cora, Citeseer and Pubmed (the performance of AdaGNN-R and AdaGNN-S is marked in \textbf{bold}).}
\label{deeper_results_rearranged1}
\begin{tabular}{c|c|cccc}
\hline
\textbf{Dataset}                      & \textbf{Model}        & \textbf{2 Layer}   & \textbf{4 Layer}   & \textbf{8 Layer}   & \textbf{16 Layer}  \\
\hline
\multirow{8}{*}{BlogCatalog} 
                             & GCN          & 73.98 $\pm$ 0.6\% & 69.71 $\pm$ 0.4\% & 37.61 $\pm$ 2.2\% & 20.61 $\pm$ 1.9\% \\
                             & GraphSAGE    & 70.41 $\pm$ 0.5\% & 67.03 $\pm$ 0.5\% & 39.15 $\pm$ 1.6\% & 18.34 $\pm$ 3.9\% \\
                             & SGC          & 73.97 $\pm$ 0.6\% & 68.94 $\pm$ 0.8\% & 47.94 $\pm$ 0.9\% & 29.02 $\pm$ 1.7\% \\
                             & DropEdge-GCN & 74.17 $\pm$ 0.7\% & 70.96 $\pm$ 1.3\% & 60.51 $\pm$ 2.4\% & 51.88 $\pm$ 0.8\% \\
                             & Pairnorm-GCN-SI      & 67.32 $\pm$ 0.7\%      & 63.61 $\pm$ 0.9\%      & 65.04 $\pm$ 0.6\%      & 67.51 $\pm$ 0.4\%     \\
                             & Pairnorm-GCN-SCS     & 71.67 $\pm$ 0.3\%      & 67.01 $\pm$ 0.2\%      & 69.30 $\pm$ 0.7\%      & 69.75 $\pm$ 1.2\%      \\
                             \cline{2-6} 
                             & \textbf{AdaGNN-R}       & \textbf{86.80 $\pm$ 0.3\%} & \textbf{87.04 $\pm$ 0.2\%} & \textbf{86.68 $\pm$ 0.1\%} & \textbf{86.44 $\pm$ 0.5\%} \\
                             & \textbf{AdaGNN-S}       & \textbf{88.50 $\pm$ 0.2\%} & \textbf{88.79 $\pm$ 0.2\%} & \textbf{88.81 $\pm$ 0.1\%} & \textbf{88.19 $\pm$ 0.2\%} \\
\hline
\multirow{8}{*}{Flickr}      
                             & GCN          & 59.82 $\pm$ 0.7\% & 32.21 $\pm$ 0.9\% & 12.20 $\pm$ 1.0\% & 13.29 $\pm$ 0.1\% \\
                             & GraphSAGE    & 54.54 $\pm$ 0.7\% & 34.72 $\pm$ 0.6\% & 11.30 $\pm$ 0.1\% & 11.20 $\pm$ 0.2\% \\
                             & SGC          & 60.74 $\pm$ 0.8\% & 41.11 $\pm$ 1.5\% & 16.64 $\pm$ 2.2\% & 14.40 $\pm$ 1.3\% \\
                             & DropEdge-GCN & 58.24 $\pm$ 3.1\% & 47.68 $\pm$ 0.8\% & 36.16 $\pm$ 0.3\% & 27.30 $\pm$ 1.6\% \\
                             & Pairnorm-GCN-SI      & 46.43 $\pm$ 0.2\%      & 39.43 $\pm$ 1.3\%      & 39.12 $\pm$ 0.8\%      & 38.24 $\pm$ 0.2\%     \\
                             & Pairnorm-GCN-SCS     & 48.93 $\pm$ 0.4\%      & 39.44 $\pm$ 1.2\%      & 34.79 $\pm$ 0.3\%      & 38.17 $\pm$ 0.2\%      \\
                             \cline{2-6} 
                             & \textbf{AdaGNN-R}       & \textbf{68.41 $\pm$ 0.5\%} & \textbf{68.29 $\pm$ 0.5\%} & \textbf{65.92 $\pm$ 0.1\%} & \textbf{66.42 $\pm$ 0.2\%} \\
                             & \textbf{AdaGNN-S}       & \textbf{71.68 $\pm$ 0.3\%} & \textbf{72.03 $\pm$ 0.3\%} & \textbf{72.93 $\pm$ 0.1\%} & \textbf{73.03 $\pm$ 0.4\%}  \\
\hline
\multirow{8}{*}{Cora}        
                             & GCN          & 81.30 $\pm$ 0.3\% & 77.48 $\pm$ 0.4\% & 65.38 $\pm$ 0.2\% & 24.28 $\pm$ 4.5\% \\
                             & GraphSAGE    & 80.75 $\pm$ 0.1\% & 78.52 $\pm$ 1.5\% & 70.26 $\pm$ 2.1\% & 26.00 $\pm$ 5.7\% \\
                             & SGC          & 79.86 $\pm$ 0.5\% & 78.50 $\pm$ 0.4\% & 71.50 $\pm$ 1.8\% & 65.60 $\pm$ 4.1\% \\
                             & DropEdge-GCN & 81.20 $\pm$ 0.4\% & 78.70 $\pm$ 0.4\% & 75.37 $\pm$ 1.6\% & 68.38 $\pm$ 1.6\% \\
                             & Pairnorm-GCN-SI      & 79.90 $\pm$ 0.5\%      & 79.76 $\pm$ 0.5\%      & 78.49 $\pm$ 0.5\%      & 74.39 $\pm$ 0.2\%     \\
                             & Pairnorm-GCN-SCS     & 81.90 $\pm$ 0.9\%      & 78.69 $\pm$ 1.1\%      & 78.18 $\pm$ 1.3\%      & 73.00 $\pm$ 0.5\%      \\
                             \cline{2-6} 
                             & \textbf{AdaGNN-R}       & \textbf{81.73 $\pm$ 0.3\%} & \textbf{81.83 $\pm$ 0.1\%} & \textbf{82.30 $\pm$ 0.4\%} & \textbf{82.18 $\pm$ 0.2\%} \\
                             & \textbf{AdaGNN-S}       & \textbf{81.60 $\pm$ 0.1\%} & \textbf{81.80 $\pm$ 0.2\%} & \textbf{82.60 $\pm$ 0.2\%} & \textbf{82.39 $\pm$ 0.3\%} \\
\hline
\multirow{8}{*}{Citeseer}    
                             & GCN          & 71.30 $\pm$ 0.3\% & 63.16 $\pm$ 1.5\% & 34.20 $\pm$ 3.0\% & 33.28 $\pm$ 5.2\% \\
                             & GraphSAGE    & 70.85 $\pm$ 0.5\% & 66.58 $\pm$ 1.4\% & 47.00 $\pm$ 4.4\% & 30.37 $\pm$ 3.7\% \\
                             & SGC          & 69.13 $\pm$ 0.2\% & 69.03 $\pm$ 0.3\% & 67.53 $\pm$ 0.1\% & 66.22 $\pm$ 0.9\% \\
                             & DropEdge-GCN & 71.20 $\pm$ 0.4\% & 66.10 $\pm$ 1.1\% & 52.38 $\pm$ 1.7\% & 49.22 $\pm$ 0.9\% \\
                             & Pairnorm-GCN-SI      & 67.71 $\pm$ 0.4\%      & 66.21 $\pm$ 0.7\%      & 64.88 $\pm$ 0.7\%      & 60.55 $\pm$ 1.6\%     \\
                             & Pairnorm-GCN-SCS     & 68.08 $\pm$ 1.4\%      & 64.56 $\pm$ 1.6\%      & 60.90 $\pm$ 1.8\%      & 56.33 $\pm$ 1.6\%      \\
                             \cline{2-6} 
                             & \textbf{AdaGNN-R}       & \textbf{70.30 $\pm$ 0.2\%} & \textbf{70.70 $\pm$ 0.2\%} & \textbf{71.54 $\pm$ 0.4\%} & \textbf{70.40 $\pm$ 0.1\%} \\
                             & \textbf{AdaGNN-S}       & \textbf{71.46 $\pm$ 0.2\%} & \textbf{71.95 $\pm$ 0.1\%} & \textbf{72.03 $\pm$ 0.1\%} & \textbf{71.34 $\pm$ 0.3\%}  \\
\hline
\multirow{8}{*}{Pubmed}      
                             & GCN          & 78.58 $\pm$ 0.6\% & 72.02 $\pm$ 0.5\% & 61.80 $\pm$ 6.7\% & 54.10 $\pm$ 8.4\% \\
                             & GraphSAGE    & 78.22 $\pm$ 0.2\% & 72.05 $\pm$ 2.0\% & 70.23 $\pm$ 4.8\% & 56.03 $\pm$ 5.7\% \\
                             & SGC          & 77.60 $\pm$ 0.4\% & 75.27 $\pm$ 0.9\% & 71.20 $\pm$ 0.3\% & 60.00 $\pm$ 2.3\% \\
                             & DropEdge-GCN & 78.33 $\pm$ 0.3\% & 77.70 $\pm$ 1.0\% & 74.80 $\pm$ 0.8\% & 71.97 $\pm$ 1.2\% \\
                             & Pairnorm-GCN-SI      & 76.80 $\pm$ 0.4\%      & 77.37 $\pm$ 0.2\%      & 78.11 $\pm$ 0.6\%      & 77.51 $\pm$ 0.9\%     \\
                             & Pairnorm-GCN-SCS     & 78.46 $\pm$ 0.1\%      & 75.65 $\pm$ 0.9\%      & 77.74 $\pm$ 1.2\%      & 71.37 $\pm$ 0.8\%      \\
                             \cline{2-6} 
                             & \textbf{AdaGNN-R}       & \textbf{78.90 $\pm$ 0.1\%} & \textbf{78.40 $\pm$ 0.1\%} & \textbf{78.50 $\pm$ 0.2\%} & \textbf{78.00 $\pm$ 0.2\%} \\
                             & \textbf{AdaGNN-S}       & \textbf{78.60 $\pm$ 0.2\%} & \textbf{78.40 $\pm$ 0.2\%} & \textbf{78.70 $\pm$ 0.2\%} & \textbf{78.60 $\pm$ 0.1\%}  \\
\hline
\end{tabular}
\end{table*}

\subsection{Experimental Settings}
\textbf{Datasets.} To comprehensively explore the performance of AdaGNN, we use six real-world attributed networks, including two social networks BlogCatalog and Flickr~\cite{li2015unsupervised}, one co-author network ACM~\cite{tang2008arnetminer}, and three citation networks Cora, Citeseer, and Pubmed~\cite{kipf2016semi}. It should be noted that the average node degree of two social networks BlogCatalog and Flickr are much higher than others.

\textbf{Baselines and evaluation protocols.} We design two different versions of AdaGNN by instantiating $\tilde{\mathbf{L}}$ as $\tilde{\mathbf{L}}_{rw}$ and $\tilde{\mathbf{L}}_{sym}$, and we name these two implementations as AdaGNN-R and AdaGNN-S. These two methods are compared with the following state-of-the-art GNNs: (1) \textit{GCN}~\cite{kipf2016semi}; (2) \textit{GraphSAGE}~\cite{hamilton2017inductive}; (3) \textit{SGC}~\cite{wu2019simplifying}. Also, we compare our framework with the recently developed methods that tackle over-smoothing: (4) \textit{DropEdge}~\cite{rong2019dropedge} -- which relieves over-smoothing issue by edge masking; (5) \textit{PairNorm}~\cite{zhao2019pairnorm} -- which tackles over-smoothing with a normalization layer and 
its two different corresponding implementations are \textit{PairNorm-SI} and \textit{PairNorm-SCS}. We use the mean aggregator and assign the sampling rate of 1 for GraphSAGE. Meanwhile, \textit{GCN} layers are used as the backbone of \textit{DropEdge}, \textit{PairNorm-SI}, and \textit{PairNorm-SCS}. All methods are compared on the semi-supervised node classification task. For BlogCatalog, Flickr and ACM datasets, we randomly sample 10\% nodes for training, 20\% for validation, and the rest 70\% for test. For Cora, Citeseer, and Pubmed, we use the same split as~\cite{kipf2016semi, yang2016revisiting}. Average classification accuracy on test dataset is presented in Table~\ref{deeper_results_rearranged2} and~\ref{deeper_results_rearranged1}, where all the results are averaged over 10 different runs.

\textbf{Implementation details.} The proposed framework AdaGNN is implemented in Pytorch~\cite{paszke2017automatic} with Adam optimizer~\cite{kingma2014adam}, and the embedding dimensions are set to be 128 across all layers except the first layer and the last layer. ReLU and Softmax activation functions are used for the first and the last layer, and the rest layers do not use any activation functions. For the baseline methods, we use their released implementations, and the hidden unit number is also specified as 128 for a fair comparison. For BlogCatalog, Flickr, Cora, Citeseer and Pubmed, we vary the number of layers in \{2, 4, 8, 16\} for all methods; for ACM, we vary it in \{2, 8, 32, 128\} for all methods to have a better observation of the over-smoothing issue. Early stopping is used for model training. To train models with 2 layers, the maximum number of epochs is 300, learning rate is 0.01, dropout rate is 0.5, $\alpha$ of $\ell_{1}$-norm (only for AdaGNN) is 1e-6, $\beta$ of $\ell_{2}$-norm (for all methods) is 9e-4. For models with deeper layers, these hyper-parameters (e.g., learning rate, dropout rate) are selected according to the best performance on the validation set.



\subsection{Experimental Results}
In this subsection, we show the detailed experimental results w.r.t. the research questions proposed above.

\textbf{Model Expressiveness (RQ1)} To validate the expressiveness of the proposed AdaGNN, we compare AdaGNN-R and AdaGNN-S with different baselines on semi-supervised node classification. We vary the model layer $K$ from 2 to 128 for ACM and  2 to 16 for other datasets, and present the performance of all models w.r.t. layer number in Table~\ref{deeper_results_rearranged2} and Table~\ref{deeper_results_rearranged1}, respectively. Here we adopt different layer settings for different datasets in order to get better observation of over-smoothing for different GNNs, and make comparison with other state-of-the-art baselines tackling over-smoothing issue. Based on the experimental performance, we make the following observations: (1) The proposed AdaGNN-R and AdaGNN-S outperform baseline methods in most cases, which demonstrates that the designed adaptive frequency response filter can indeed increase the fitting capability of the model by learning more discriminative embeddings. (2) The performance improvements of AdaGNN-R and AdaGNN-S are more obvious on BlogCatalog and Flickr compared with other datasets. The reason could be attributed to their high average node degree of social network (as indicated in Table \ref{tb:Statistics}) -- nodes are influenced by more neighbors during neighborhood aggregation, which is consistent with the observations in previous literature~\cite{chen2019measuring}. For such dataset with high average node degree, the adaptive frequency response provided by AdaGNN can help achieve more appropriate feature smoothness. This provide us larger performance improvement compared with other GNNs.

\textbf{Over-smoothing (RQ2)} Now we answer RQ2 by investigating how models perform when the layer number increases. We have the following observations based on Table~\ref{deeper_results_rearranged2} and Table~\ref{deeper_results_rearranged1}: (1) The best performance is achieved at shallow layers for all baselines; however the performance of conventional GNNs (e.g., GCN, GraphSAGE, and SGC) drops sharply when layer goes deeper, revealing that they suffer from the over-smoothing issue at deeper layers. (2) DropEdge and PairNorms (including PairNorm-SI and -SCS) are recently proposed state-of-the-art methods to relieve over-smoothness, whose performance does not drop as fast as conventional GNNs. In particular, the performance of DropEdge is comparable to its backbone GCN while PairNorms is inferior to its backbone in many scenarios, which is consistent with the observations in the original papers~\cite{zhao2019pairnorm}. (3) The performances of the proposed AdaGNN-R and AdaGNN-S are further improved with more powerful representations at deeper layers in our proposed framework, demonstrating the effectiveness of the proposed filter in tackling over-smoothing and extracting more information from deeper layers.

\begin{figure*}[!htbp]
\vspace{-3mm}
	\centering 
	\subfloat[Frequency response.]{
        \includegraphics[width=0.4\textwidth]{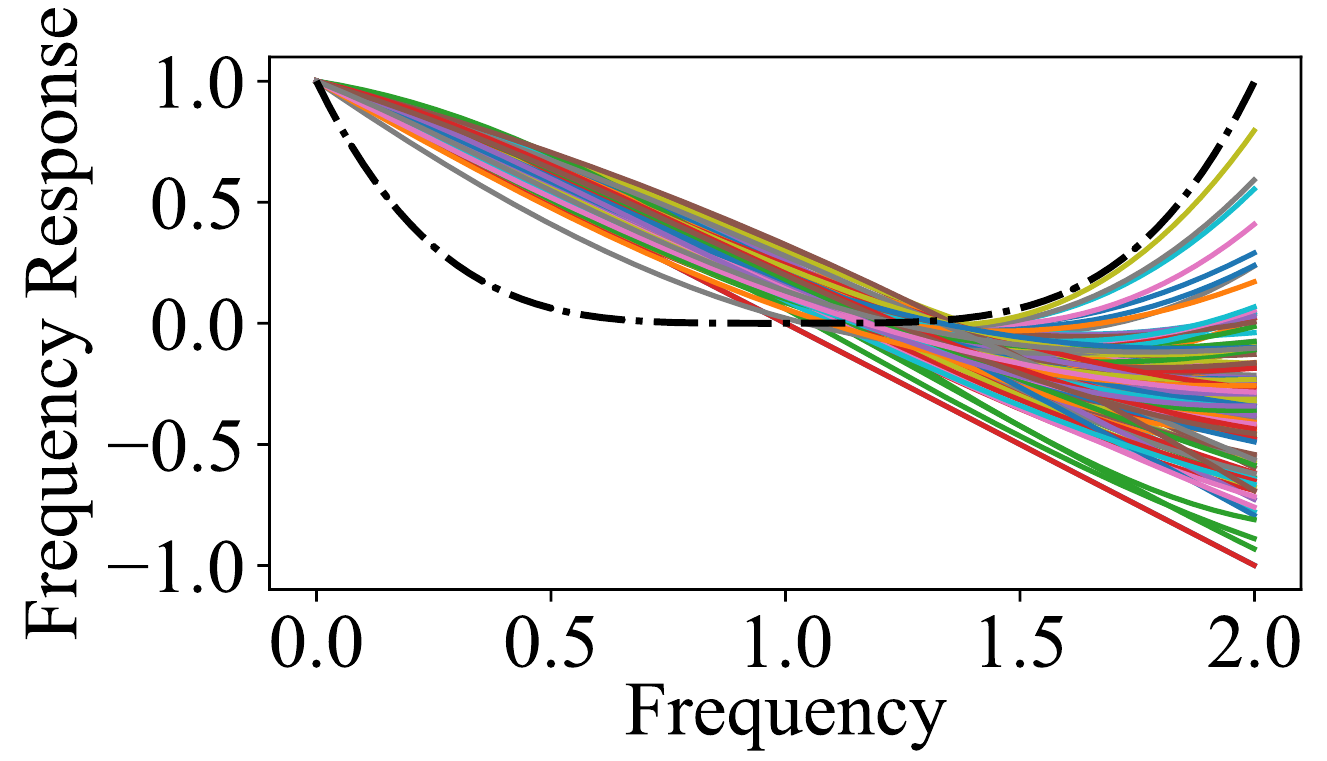}
        \label{fig:filter}
    } 
    \hspace{10mm}
    \subfloat[Model ablation study.]{
        \includegraphics[width=0.4\textwidth]{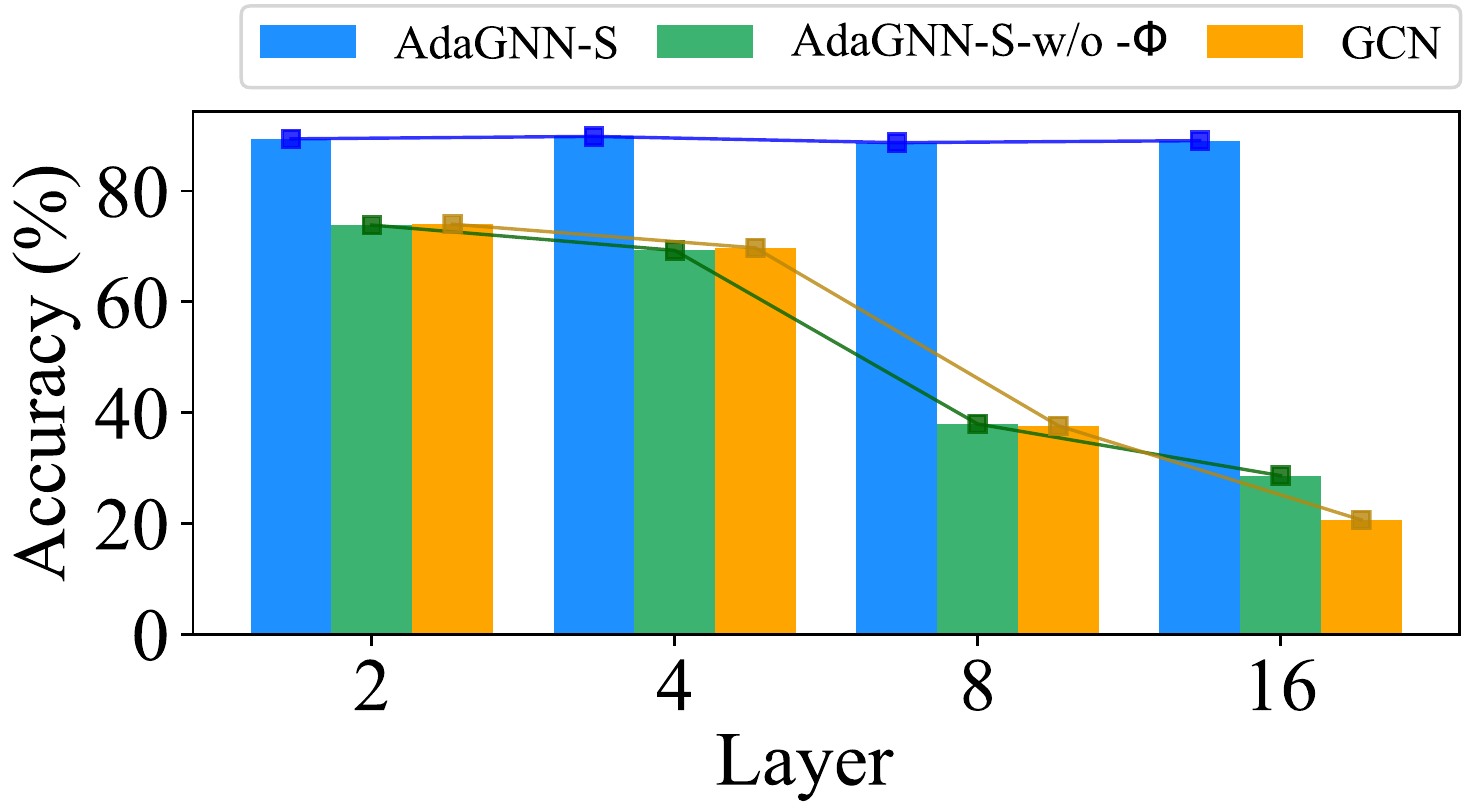}
        \label{fig:ablation}
    }
\vspace{-2mm}
	\caption{ (a) Frequency response function of 4-layered AdaGNN-S and SGC on Flickr. The black dashed curve denotes the response of SGC while other solid curves denote the responses of AdaGNN-S across feature channels. (b) Model ablation study of AdaGNN-S on BlogCatalog, where layer number varies from \{2, 4, 8, 16\}.}  
	\vspace{-2mm}
	\label{fig:filter&ablation}   
\end{figure*}

\textbf{Filter Analysis (RQ3)} To answer RQ3, in Fig.~\ref{fig:filter}, we provide a detailed visualized comparison between the fixed filter of SGC and the learned filters of AdaGNN-S for different feature channels. It should be noted that the frequency response of the two models can be compared following the order of eigenvalues due to both AdaGNN and SGC are graph Laplacian matrix based filters~\cite{chung1997spectral}. 
Frequency response from 0 (the lowest frequency for undirected graph based on Laplacian spectrum) to 2 (the highest frequency for undirected graph based on Laplacian spectrum~\cite{chung1997spectral}) is presented without cut-off for generalization purpose. As can be shown, firstly, the frequency response of AdaGNN is naturally enforced with different characteristics for different feature channel after the optimization. This demonstrate the effectiveness of the learnable filter in our proposed AdaGNN. Secondly, the learned frequency response function of AdaGNN-S varies across feature channels while the function of SGC treats these channels equally. Compared with the band-stop frequency response function of SGC, AdaGNN-S preserves more middle-frequency components, which could help to learn more discriminative representations. Finally, the response of AdaGNN-S is highly selective across high-frequency components, revealing that some of them are complementary to low/middle frequencies to improve performance while some can be taken as noise.

\textbf{Ablation Study (RQ4)} Now we perform ablation study to answer RQ4. It shold be noted that AdaGNN-S reduces to GCN when we remove the parameter matrix $\mathbf{\Phi}$ and incorporate the weight matrix and activation function in the intermedia layer. Consequently, in order to individually explore the contribution of the two components to the final performance, we compare the performance of three variants of AdaGNN for ablation study, i.e., AdaGNN-S, AdaGNN-S-w/o-$\mathbf{\Phi}$, and GCN in Fig.~\ref{fig:ablation}. We make the discussion as follows. Firstly, it can be clearly observed from Fig.~\ref{fig:ablation} that the learning performance is greatly reduced when $\mathbf{\Phi}$ is removed at each layer. This demonstrates the indispensable contribution of the learnable filter to the shown performance improvement on node classification. At the same time, compared with original AdaGNN-S, AdaGNN-S-w/o-$\mathbf{\Phi}$ shows the tendency of over-smoothing in deeper layers, which can also be observed on GCN. This indicates that the learnable diagonal matrix $\mathbf{\Phi}$ in AdaGNN-S also contributes to the over-smoothing relief when more layers are stacked together.


\section{Relatex Work}

\textbf{Spectral-based Graph Neural Networks.} Existing graph neural network models are often categorized as spectral-based and spatial-based methods depending on the operation domain~\cite{wu2020comprehensive,zhang2019graph}. Graph Signal Processing lays a solid mathematical foundation for spectral-based methods by enabling them to define graph filter in the spectral domain. Bruna et al.~\cite{bruna2013spectral} first proposed to generalize the convolution operations in CNN to graphs and define graph filter with the spectrum of the graph Laplacian matrix. Later on, Defferrard et al.~\cite{defferrard2016convolutional} proposed a fast localized convolutional filter ChebNet based on Chebyshev polynomial which avoids expensive eigendecomposition operation and is considered as a special case of CayleyNet that applies Cayley polynomials~\cite{levie2018cayleynets}. The seminal work of GCN~\cite{kipf2016semi} utilizes a spectral filter by truncating Chebyshev polynomial to only first order and the filter can be regarded as neighborhood aggregation in the spatial domain. SGC~\cite{wu2019simplifying} further simplifies GCN by collapsing weight matrix in consecutive layers and showed that there are redundant computations in GCN. Additionally, recent efforts attempt to improve the spectral filter from different perspectives, such as learning hidden structural relations~\cite{li2018adaptive}, emphasizing both low and high frequencies~\cite{bo2021beyond} and capturing both local and global information~\cite{zhuang2018dual}. Despite their empirical effectiveness, these attempts cannot well characterize the varying importance of frequencies for learning. Different from all previous works, in this paper, we achieve a learnable filter individually for each feature dimension. In this way, AdaGNN achieves natural over-smoothing alleviation with more discriminatove representations, which also contributes to the performance.

\textbf{Spatial-based Graph Neural Networks.} Spatial-based methods perform convolution in the spatial domain by aggregating and transforming the information of neighboring nodes. Different methods in this family mainly differ in the way how the aggregation function is designed. One of the earliest attempts NN4G~\cite{micheli2009neural} sums up the information from a central node's neighbors. DCNN~\cite{atwood2016diffusion} regards graph convolutions as a diffusion process w.r.t. specific probabilities to attain equilibrium after several rounds. To better distinguish the importance of different neighbors for information aggregation, attention mechanism is also utilized in GAT~\cite{velivckovic2017graph}. MPNN~\cite{gilmer2017neural} generalizes different spatial-based methods as a unified message-passing framework. GraphSAGE~\cite{hamilton2017inductive} aggregates neighborhood information via mean/max/LSTM pooling. GIN~\cite{xu2018powerful} allocates a learnable parameter for the center node when performing information aggregation, which empowers the model stronger capability to differentiate different graph structures. More recently, a myriad of more  sophisticated aggregation strategies compared with previous works are developed and a more detailed review can be referred to~\cite{wu2020comprehensive,zhang2020deep}.

\textbf{Over-smoothing of Graph Neural Networks.} Information aggregation from node neighbors is a critical step of GNNs, which smoothes node feature representations over the whole graph~\cite{zhu2003semi,zhou2004learning,belkin2006manifold}. In the spectral domain, this can be interpreted as weakening the high-frequency components of input signals. For example, studies have shown that the convolution operation in GCN corresponds to a single fixed low-pass filter~\cite{nt2019revisiting}. When multiple convolution layers are stacked and the model goes deeper, the over-smoothing issue is inevitable, which means node representations converge to similar values. Li et al.~\cite{li2018deeper} proved that GCN is actually a kind of Laplacian smoothing process, and proposed the challenge of over-smoothing for the first time. After that, some studies demonstrate that certain level of smoothness benefits node representation learning while over-smoothing broadly exists in deeper GNNs~\cite{deng2019batch, chen2019measuring}. More recently, some researches attempt to relieve this problem via residual-like connections~\cite{li2019recursive, chen2020simple, liu2020towards}. Nevertheless, such methods are unable to avoid the situation where a node is overwhelmed in the information of its neighbors in the information aggregation process. There are also works directly relieve over-smoothing in this process via using either edge masking~\cite{rong2019dropedge} or re-normalization~\cite{zhao2019pairnorm}; however performance still obviously reduces when models go deeper. Consequently, it remains a challenging problem to directly relieve over-smoothing in the information propagating process. To the best of our knowledge, we are the first to provide an understanding of this problem from the perspective of the spectral filter, and our experiments also demonstrate its superiority over other prevalent solutions such as DropEdge~\cite{rong2019dropedge} and PairNorm~\cite{zhao2019pairnorm}.

\section{Conclusion}
Existing spectral GNNs mainly apply fixed filters for the convolution operation, where such non-learnable filter leads to two problems. Firstly, due to that the graph filter is fixed, their expressiveness is limited in the learning process; secondly, it could be hard for such GNNs to achieve an appropriate level of feature smoothness, and over-smoothing happens unavoidbly in deeper layers. To tackle the above mentioned problems, in this paper, we propose a novel framework AdaGNN with an adaptive frequency response filter. By learning to individually control information flow for different feature channels, the proposed filter is able to adaptively adjust the importance of different frequency components of each input feature channel, which leads to a learnable filter when multiple layers are stacked together. AdaGNN also learns more discriminative representations via achieving different levels of smoothness for different feature channels. We provide theoretical analysis for the proposed AdaGNN from different aspects, and empirical experimental evaluations also demonstrate its superiority on performance over state-of-the-art GNNs and over-smoothing alleviation over other state-of-the-art baselines. We will leave the fairness issues of the proposed AdaGNN framework as our future research directions.

\section{Acknowledgements}
This material is, in part, supported by the National Science Foundation (NSF) under grant number 2006844 and 2006861. We would like to thank the anonymous reviewers for their constructive feedback.

\clearpage

\bibliographystyle{ACM-Reference-Format}
\bibliography{reference.bib}

\end{document}